\def\year{2017}\relax
\documentclass[letterpaper]{article}
\usepackage{aaai17}
\usepackage{times}
\usepackage{helvet}
\usepackage{courier}

\usepackage{amsfonts}
\usepackage{amsmath}
\newtheorem{theorem}{Theorem}
\usepackage{cases}
\usepackage{multirow}
\usepackage{subfigure}
\usepackage{graphicx}

\begin{document}
%
\title{Self-Paced Multi-Task Learning}
\author{Changsheng Li$^{12}$, Junchi Yan$^{12}$\thanks{Contributed equally. Junchi Yan is the corresponding author.}, Fan Wei$^{3}$$^{*}$, Weishan Dong$^{2}$, Qingshan Liu$^{4}$, Hongyuan Zha$^{51}$\\
$^1$East China Normal University \quad $^2$IBM Research -- China\\
$^3$Stanford University \quad  $^4$Nanjing University of Info. Science $\&$ Tech \quad $^5$Georgia Institute of Technology\\
\{lcsheng,dongweis\}@cn.ibm.com, \{jcyan,zha\}@sei.ecnu.edu.cn, fanwei@stanford.edu, qsliu@nuist.edu.cn}
\maketitle
\begin{abstract}
Multi-task learning is a paradigm, where multiple tasks are jointly learnt.
Previous multi-task learning models usually treat all tasks and instances per task equally during learning.
Inspired by the fact that humans often learn from easy concepts to hard ones in the cognitive process, in this paper, we propose a novel multi-task learning framework that attempts to learn the tasks by simultaneously taking into consideration the complexities of both tasks and instances per task.
We propose a novel formulation by presenting a new task-oriented regularizer that can jointly prioritize tasks and instances.
Thus it can be interpreted as a self-paced learner for multi-task learning.
An efficient block coordinate descent algorithm is developed to solve the proposed objective function, and the convergence of the algorithm can be guaranteed.
Experimental results on the toy and real-world datasets demonstrate the effectiveness of the proposed approach, compared to the state-of-the-arts.
\end{abstract}

\section{Introduction}
The paradigm of multi-task learning (MTL) involves learning several prediction tasks simultaneously.
One basic assumption in MTL is that there exists common or related information among tasks, and learning such information can result in better prediction performance than learning each task independently \cite{caruana1997multitask}.
It is particularly desirable to share such information across tasks, when there are many related tasks but the available training data are limited.
Due to its empirical success and good theoretical foundations, MTL has been applied to various domains, including disease modeling and prediction \cite{zhou2011multi},  web image and video search \cite{wang2009boosted}, and relative attributes learning \cite{chen2014predicting}.

Many MTL methods have been proposed, which in general can be categorized into two classes based on the principal way to learn the relatedness \cite{kang2011learning,pu2013multiple,PuNC16,ZhongNC16}.
The first class assumes that \emph{all} the tasks share common yet low-rank feature representations \cite{argyriou2008convex,zhang2010probabilistic,yang2014salient,kim2010tree},
and the other class of methods assumes that the model parameters used by the tasks are related to each other \cite{schwaighofer2004learning,ando2005framework,yang2016large,zhang2010transfer}.
In these methods, the assumption that common information is shared across {all} tasks is strong in certain cases. Thus recent methods propose to group tasks or detect outlier tasks, which assume that there exists common information only within a \emph{subset} of tasks, or exist outlier tasks having no relation with other tasks \cite{jalali2010dirty,kumar2012learning}.
However, when learning the related information across tasks, the algorithms above treat all tasks equally and all instances per task equally, in other words, there is no mechanism to control the order of the tasks and the instances to learn among these methods.

Different from previous methods, in this paper, we propose a novel MTL framework by simultaneously taking into consideration the complexities of both instances and tasks during learning. This idea is inspired by the fact that humans often learn from easy concepts to hard ones in the cognitive process \cite{elman1993learning,bengio2009curriculum}. For example, a student often starts with easier concepts (e.g. recognizing objects in simple scenes where an object is clearly visible) and builds up to more complex ones (e.g. cluttered images with occlusions).
Such a learning process is inherently essential for human education and cognition.
Similarly, in the regime of MTL, not only do there exist `easy' to `hard' instances, but also `easy' to `hard' tasks.
For instance, recognizing monkeys from the image set consisting of monkeys and tigers is a relatively `easy' task, while recognizing baboons from the image set consisting of baboons and orangutans is a relatively `hard' task. In the first task, an image of monkey with plain background is a relatively `easy' positive instance, while one with complex background is relatively `hard' positive.
If a multi-task learner can learn the related information among tasks by first using `easy' tasks and instances and then gradually involving `hard' ones, as human brain does, then it can benefit more with less effort.

We name the proposed MTL framework, \emph{\underline{S}elf-\underline{P}aced \underline{M}ulti-\underline{T}ask \underline{L}earning} (SPMTL), which aims to learn the multi-task model in a self-paced regime.
The contributions of this paper are threefold:
\begin{itemize}
\item
It is the first work, to our best knowledge, where a principled MTL model jointly takes into consideration the complexities of both training instances and tasks.
Our model can be interpreted as a self-paced MTL model to explore common information among tasks.
\item
We propose a new regularizer, which can set priorities for both tasks and instances in each iteration, and use smooth weights for such priorities.
To the best of our knowledge, this is also the first task-oriented self-paced regularizer tailored to MTL in literature.
\item
An efficient block coordinate descent algorithm is developed to solve the proposed objective function, and the convergence of the algorithm can be guaranteed. Experimental results on the toy and real-world datasets demonstrate the effectiveness of the proposed approach.
\end{itemize}

\section{Related Work}
Multi-task learning (MTL) aims to learn the related information across tasks, so as to improve the prediction performance of the model.
However, most of the existing multi-task models learn such information by treating  all tasks and instances equally.
Recently, an \emph{active} \emph{online} MTL method \cite{ruvolo2013active} is proposed, which can actively select the next task to learn, so as to maximize prediction performance on future learning tasks. In addition, two task selection algorithms \cite{ban2014smart} are also proposed for \emph{active online} MTL, which are based on the QR-decomposition and minimal-loss principles, respectively. Although these two methods consider the order of the tasks during training, but they do not adopt the strategy learning from `easy' tasks to `hard' tasks.

More recently,  a novel task selection method \cite{pentina2014curriculum} based on curriculum learning \cite{bengio2009curriculum} is proposed  for \emph{batch} MTL.
The method aims to solve tasks in a sequential manner by transferring information from a previously learned task to the next one instead of solving all of them simultaneously.
However, this method transfers information unidirectionally, i.e., once one task is learned, it will be not affected by the subsequent tasks to learn. In a dynamic and complex learning process of multi-task model, such an information propagation way may be not optimal.
In addition, this method ignores the `easiness' and `hardness' properties of instances.

Recently, a new learning regime, called self-paced learning (SPL) \cite{kumar2010self}, is proposed for several learning problems \cite{zhang2015self,xu2015multi}. Different from curriculum learning usually designing curriculums based on certain heuristical `easiness' measurements, SPL can automatically and dynamically choose the order in which training instances are processed for solving a non-convex learning problem \cite{jiang2014self}. Although SPL has been studied for single task learning \cite{kumar2010self,jiang2014self}, there has been no effort put on MTL until now.

\section{Self-Paced Multi-Task Learning}
Suppose we are given $m$ learning tasks $\{T_i\}^m_{i=1}$.
For the $i$-th task $T_i$, the training set $\mathcal{D}_i$ consists of $n_i$ data points $\{(\mathbf{x}_{ij}, {y}_{ij})\}_{j=1}^{n_i}$ , where $\mathbf{x}_{ij}\in \mathbb{R}^d$ is the feature representation of the $j$-th instance and $y_{ij}$ is its corresponding output, such as $y_{ij}\in\mathbb{R}$ for regression and $y_{ij}\in \{-1,1\}$ for binary classification problem.
The total number of the training instances is $n=\sum_{i=1}^mn_i$.
The prediction model for the $i$-th task is defined as $g(\mathbf{p}_i,\mathbf{x}_{ij})=\mathbf{p}_i^T\mathbf{x}_{ij}$.
Generally speaking, the objective of multi-task learning (MTL) is to derive optimal prediction models for all $m$ tasks simultaneously.
Inspired by the fact that humans often learn concepts from the easiest to the hardest, we incorporate the easy-to-hard strategy operated on tasks and instances simultaneously into the learning process of MTL.
Thus, we propose a new objective function:
\begin{align}\label{obj1}
\min\limits_{\mathbf{w,U,V}} &\sum_{i=1}^m\frac{1}{n_i}\sum_{j=1}^{n_i}w_{j}^{(i)}\mathcal{L}(y_{ij},\mathbf{v}_i^T\mathbf{U}^T\mathbf{x}_{ij})+\alpha\|\mathbf{U}\|_F^2\nonumber\\
& \ \ +\beta\|\mathbf{V}\|_1+f(\mathbf{w},\lambda,\gamma)\\
&s.t. \ w_{j}^{(i)}\in [0,1], \forall j=1,\ldots n_i, i=1,\ldots, m,  \nonumber
\end{align}
where $\mathbf{V}=[\mathbf{v}_1,\ldots,\mathbf{v}_m]\in\mathbf{R}^{k\times m}$. $\mathbf{w}=[w_1^{(1)},\ldots,w_{n_1}^{(1)},w_1^{(2)},\ldots,w_{n_2}^{(2)},\ldots,w_{n_m}^{(m)}]\in \mathbb{R}^n$ denotes the importance weights imposed on all the instances.
$f(\mathbf{w},\lambda,\gamma)$ denotes the self-paced regularizer that \emph{dynamically} determines which instances and tasks used for training.
$\mathcal{L}(y_{ij},\mathbf{v}_i^T\mathbf{U}^T\mathbf{x}_{ij})$ is the empirical loss on the training data points $({\mathbf{x}_{ij}},y_{ij})$.
$\mathbf{U}$ is a $d\times k$ matrix with each column representing a basis. $\mathbf{V}$ is a $k \times m$ matrix whose columns contain the coefficients of the linear combination of the basis for the corresponding tasks. $\alpha\geq0$ and $\beta\geq0$ are two trade-off parameters.

Next, let us have a closer look at the objective function (\ref{obj1}). Different from the traditional empirical loss on the training data, the first term in (\ref{obj1}) is a weighted loss term on all the training instances and tasks.
The second term is used to control the complexity of $\mathbf{U}$, and the third term aims to make $\mathbf{V}$ sparse.
In (\ref{obj1}), we assume that the weight vector $\mathbf{p}_i$ of each task can be represented as a linear combination of a subset of $k$ basis tasks, i.e., $\mathbf{p}_i=\mathbf{U}\mathbf{v}_i$. Since we expect $\mathbf{v}_i$ is sparse, a subset of $k$ basis tasks is used for representing the weight vector $\mathbf{p}_i$.
By this means, the tasks with the same basis can be seen as belonging to the same group, while the tasks whose basis are orthogonal are sure to belong to different groups. The partial overlapping of bases enables the algorithm to model those tasks which are not in the same group but still share some common information.
The last term is our proposed self-paced regularizer to control which tasks and instances first to be involved in the learning process, and which ones to be gradually taken into consideration. Next, we will introduce the last term in detail.

In order to simultaneously perform the easy-to-hard strategy on both instances and tasks, we propose a new self-paced regularizer defined as:
\begin{align}\label{obj3}
f(\mathbf{w},\lambda,\gamma)&=-\lambda\sum_{i=1}^m\sum_{j=1}^{n_i}w_{j}^{(i)}\!+\!\gamma\sum_{i=1}^m\sqrt{\frac{1}{n_i}\sum_{j=1}^{n_i}(w^{(i)}_j)^2}\nonumber\\
&=-\lambda\sum_{i=1}^m\|\mathbf{w}^{(i)}\|_1\!+\!\gamma\sum_{i=1}^m\frac{\|\mathbf{w}^{(i)}\|_2}{\sqrt{n_i}},
\end{align}
where $\mathbf{w}^{(i)}=[w_1^{(i)},\ldots,w_{n_i}^{(i)}]\in [0,1]^{n_i}$, and thus $\mathbf{w}=[\mathbf{w}^{(1)},\ldots,\mathbf{w}^{(m)}]$.
$\lambda$ and $\gamma$ are two self-paced parameters to control the learning pace on instances and tasks.

There are two terms in Eq. (\ref{obj3}):
The first term is the negative $l_1$-norm, which favors selecting the easy instances to the hard ones per task.
Combining this term  with (\ref{obj1}), we can know that when the empirical loss $\mathcal{L}$ on the training data point $(\mathbf{x}_{ij}, y_{ij})$ is small, the weight $w_j^{(i)}$ tends to be high.
Thus this optimization process fits the intuitive concept of starting with the simplest instances (having low empirical error) well.
When gradually increasing $\lambda$ as the learning proceeds, the weights will generally become increasingly higher.
This can gradually involve harder instances for training.
The second term is an adaptive $l_{2,1}$-norm of a matrix, which favors selecting the easy tasks to the hard ones.
We use $\frac{1}{\sqrt{n_i}}$ in the second term to avoid task imbalance, when one task has so many data points that it dominates the norm.
As we know, minimizing the $l_{2,1}$ norm of a matrix can make the matrix sparse in rows or columns \cite{argyriou2008convex} in contrast to the $l_{1}$ sparsity e.g. \cite{YanSPL10,YanVCIP11}.
When combining this term with (\ref{obj1}), minimizing them will make the $\mathbf{w}^{(i)}$'s corresponding to large empirical loss $\mathcal{L}$ (i.e., hard tasks) be close to or equal to zero vectors.
In other words, this group-sparsity representation is expected to select the easiest tasks at the beginning of learning.
By gradually reducing $\gamma$, this group sparsity will become weaker, thus harder tasks will be gradually involved for training.
In the later experiment, we demonstrate that when the loss on the task level is high (hard task), group sparsity will make the weight of the task be small, i.e., this task will be not selected.

Plugging (\ref{obj3}) into (\ref{obj1}), we obtain the final objective function:
\begin{align}\label{obj4}
\min\limits_{\mathbf{w,U,V}} &\sum_{i=1}^m\frac{1}{n_i}\mathbf{w}^{(i)}\widehat{\mathcal{L}}^{(i)}+\alpha\|\mathbf{U}\|_F^2+\beta\|\mathbf{V}\|_1\nonumber\\
&\ \ -\lambda\sum_{i=1}^m\|\mathbf{w}^{(i)}\|_1+\gamma\sum_{i=1}^m\frac{\|\mathbf{w}^{(i)}\|_2 }{\sqrt{n_i}} \\
&s.t. \ \mathbf{w}^{(i)}\in [0,1]^{n_i}, \forall  i=1,\ldots, m, \nonumber
\end{align}
where the vector $\widehat{\mathcal{L}}^{(i)}=[\mathcal{L}_1^{(i)},\ldots,\mathcal{L}_{n_i}^{(i)}]^T$.
In this paper, we focus on regression tasks, and define $\mathcal{L}_j^{(i)}=\mathcal{L}(y_{ij},\mathbf{v}_i^T\mathbf{U}^T\mathbf{x}_{ij})=(y_{ij}-\mathbf{v}_i^T\mathbf{U}^T\mathbf{x}_{ij})^2$.
Note that our method can be naturally applied to classification tasks by adopting a classification loss function.

\subsection{Discussion}
In this section, we discuss the relation or differences between our model and some previously proposed methods:

The method in \cite{pentina2014curriculum} aims to propagate information unidirectionally, i.e., the information from the learned tasks will be transferred to the subsequent tasks to learn, while the information from the unlearned tasks will be not propagated back into the learned tasks.
Different from them, our method can jointly learn the model using all the selected tasks and the selected instances as the learning proceeds.
Since it depends on the current learner that a task or an instance is `easy' or `hard', the current `easy' and `hard' tasks may change when the learner is updated. Thus it is necessary to re-evaluate all tasks and instances once the learner is updated, such that the dynamic and complex learning process can be well fitted. The results in the experiment part also demonstrate that our method is better than \cite{pentina2014curriculum}.

The task-oriented self-paced regularizer proposed in this paper is motivated by SPLD \cite{jiang2014self}. SPLD aims to select the training instances from the view of both easiness and diversity, but it does not consider the order of tasks at all. Thus directly applying the regularizer of SPLD is not optimal for MTL.
Differently, our task-oriented regularizer can reach the goal that only several easy tasks are selected for training in the beginning and hard tasks are gradually involved.
Therefore, our regularizer is tailored to MTL.

GO-MTL \cite{kumar2012learning} is a task grouping method that assumes model parameters in the same group lying in a low-dimensional subspace, and allows the tasks from different groups to have overlapping information in common.
However, GO-MTL learns model parameters using all tasks and instances simultaneously without considering their orders during training.
When setting $\lambda=0$, $\gamma=0$, and $\mathbf{w}=\mathbf{1}$ in (\ref{obj4}), our method is reduced to GO-MTL.

\section{Optimization}
In this section, we discuss how to solve problem (\ref{obj4}). The objective function in (\ref{obj4}) is non-convex, so it is difficult to find the global optimal solution. We develop a block coordinate descent method to solve (\ref{obj4}), and can guarantee the convergence of the algorithm.

For solving block $\mathbf{w}_{t+1}$ with fixed blocks $\mathbf{U}_t$ and $\mathbf{V}_t$, the optimization problem can be formulated as $m$ individual problems for $m$ tasks respectively. For the $i$-th task $T_i$, the objective function becomes:
\begin{align}\label{obj5}
 \min\limits_{\mathbf{w}^{(i)}\in [0,1]^{n_i}} \frac{1}{n_i}\mathbf{w}^{(i)}\widehat{\mathcal{L}}_t^{(i)}-\lambda\|\mathbf{w}^{(i)}\|_1
 +\frac{\gamma}{\sqrt{n_i}}\|\mathbf{w}^{(i)}\|_2.
\end{align}
In order to solve (\ref{obj5}), we first assume
$\mathcal{L}^{(i)}_{1,t} \leq \mathcal{L}^{(i)}_{2,t} \leq \dots \leq \mathcal{L}^{(i)}_{n_i,t}$.
Let $p^{(i)}_t=\sum\limits_{k_0 < j < k_1} (\lambda - \frac{\mathcal{L}^{(i)}_{j,t}}{n_i})^2$, and $q^{(i)}_t=\sum\limits_{k_0 < j < k_1} (\lambda - \frac{\mathcal{L}^{(i)}_{j,t}}{n_i}).$
For each $i$ and arbitrary $k_1 > k_0$, we define $c^*_t(k_0, k_1)$, $L_t(k_0, k_1)$, $G^*_{i,t}, S^*_{i,t}$ for later computation:
\begin{enumerate}
\item
\begin{equation}
c^*_t(k_0, k_1)\!=\!
\begin{cases}
\sqrt{ k_0n_i/ (\gamma^2-n_ip_t^{(i)})}, \ \ if \frac{\gamma^2}{n_i} \neq p_t^{(i)}\nonumber\\
\left(\lambda \!-\! {\mathcal{L}^{(i)}_{k_0+1,t}}/{n_i}\right)^{-1}\!\!\!, if \frac{\gamma^2}{n_i} \!=\! p_t^{(i)}, \frac{\gamma^2}{n_i} \!<\! q_t^{(i)}\nonumber\\
0, \qquad  if \frac{\gamma^2}{n_i} \!=\! p_t^{(i)},\  \frac{\gamma^2}{n_i} \geq q_t^{(i)}.\nonumber
\end{cases}
\end{equation}
\item
$ L_t(k_0, k_1) =\sum_{j = 1}^{k_0} \frac{\mathcal{L}^{(i)}_{j,t}}{n_i}
- \lambda (k_0 + c_t^*(k_0, k_1) q_t^{(i)})
+ \frac{\gamma}{\sqrt{n_i}} \sqrt{  k_0 + c_t^*(k_0, k_1)^2 p_t^{(i)} }$.
\item $G^*_{i,t}$ be the smallest $j$ such that $\mathcal{L}^{(i)}_{j,t} \geq \lambda n_i$.
\item $S^*_{i,t}$ be the largest $j$ such that $\mathcal{L}^{(i)}_{j,t} \leq n_i\lambda - \sqrt{n_i}\gamma$.
\end{enumerate}
The following theorem gives the global optimum of (\ref{obj5}) (see the proof in supplementary materials).
\begin{theorem}
Let $k_1 = G^*_{i,t}$, and $k_0$ be obtained by optimizing the following objective function:
\begin{equation}
k_0 =  \arg \min_{S^*_{i,t} \leq k_0 <k_1} L_t(k_0, k_1)
\end{equation}
\begin{numcases}{s.t.\!\!}
\frac{\gamma^2}{n_i} -p_t^{(i)}\geq0, \  \text{or} \  \frac{\gamma^2}{n_i} -p_t^{(i)}>0\  if \ k_0 > 0  \nonumber \\
c_t^*(k_0, k_1)(\lambda \!-\! \mathcal{L}^{(i)}_{k_0+1,t}/n_i) \!<\! 1, \ \text{ if } k_0+1 \!<\! k_1  \nonumber \\
\!\!\frac{\mathcal{L}^{(i)}_{k_0,t}}{n_i}  \!\!+\!\! \frac{\gamma }{\sqrt{n_i}}\!\!\left(k_0 \!\!+\!\! {c_t^*(k_0,k_1)}^2 p_t^{(i)}\right)^{-\frac{1}{2}}\!\leq\! \lambda, \text{if } k_0\!\!+\!\!1 \!<\!k_1 \nonumber.
\end{numcases}
Then, the optimal $\mathbf{w}_{t+1}^{(i)}$ is given by,
\begin{equation}
{w}^{(i)}_{j,t+1}=
\begin{cases}
1,& if \ j \leq k_0,\\
0,& if \ j \geq k_1,\\
c_t^*(k_0, k_1)(\lambda -\frac{ \mathcal{L}^{(i)}_{j,t}}{n_i}),&if \  k_0 < j < k_1.\\
\end{cases}
\end{equation}
Thus it takes only linear time $O(n_i)$ to compute $\mathbf{w}_{t+1}^{(i)}$.
\end{theorem}


For solving $\mathbf{U}_{t+1}$ with fixed $\mathbf{w}_{t+1}$ and $\mathbf{V}_t$, the optimization problem is formulated as:
\begin{align}\label{objU}
\mathbf{U}_{t+1}=\arg\min\limits_{\mathbf{U}} \sum_{i=1}^m\frac{1}{n_i}\mathbf{w}^{(i)}_{t+1}\widehat{\mathcal{L}}_t^{(i)}+\alpha\|\mathbf{U}\|_F^2.
\end{align}
The necessary optimality condition is that the derivative of (\ref{objU}) with respective to $\mathbf{U}$ is zeros. Thus, we have
\begin{align}\label{optu}
&\sum_{i=1}^m\sum_{j=1}^{n_i}\frac{w_{j,t+1}^{(i)}}{n_i}\mathbf{x}_{ij}\mathbf{x}_{ij}^T\mathbf{U}\mathbf{v}_{i,t}\mathbf{v}^T_{i,t}+\alpha\mathbf{U}\nonumber\\
=&\sum_{i=1}^m\sum_{j=1}^{n_i}\frac{w_{j,t+1}^{(i)}}{n_i}y_{ij}\mathbf{x}_{ij}\mathbf{v}_{i,t}^T\nonumber\\
\Rightarrow & (\sum_{i=1}^m\sum_{j=1}^{n_i}\frac{w_{j,t+1}^{(i)}}{n_i}(\mathbf{v}_{i,t}\mathbf{v}^T_{i,t})\otimes(\mathbf{x}_{ij}\mathbf{x}_{ij}^T)+\alpha\mathbf{I})\text{vec}(\mathbf{U})\nonumber\\
=&\sum_{i=1}^m\sum_{j=1}^{n_i}\frac{w_{j,t+1}^{(i)}}{n_i}y_{ij}\text{vec}(\mathbf{x}_{ij}\mathbf{v}_{i,t}^T),
\end{align}
where $\otimes$ denotes the Kronecker product and $\text{vec}(\cdot)$ is an operator that reshapes a $d\times k$ matrix into a $dk\times 1$ vector.
\begin{table}
\begin{center}
\label{spmtl}
\begin{tabular}{l}
\hline
\textbf{Algorithm 1} \ Self-Paced Multi-Task Learning (SPMTL) \\
\hline
\textbf{Input:} Data matrix $\{\mathcal{D}_i\}_{i=1}^m$, number of latent tasks $k$, \\
 \ \ \ \ \ \ \ \ \ \ \ \  regularization parameters $\alpha$ and $\beta$,\\
 \ \ \ \ \ \ \ \ \ \ \ \ iterations $T_{max}$, and tolerance $\varepsilon$, $\mu_1=\mu_2>1$;\\
1. Initialize $\!\mathbf{P}\!=\!\![\mathbf{p}_1,\!\ldots, \!\mathbf{p}_m]\!$ by standard ridge regression;\\
2. Initialize $\mathbf{U}_0$ using top-$k$  singular vectors of $\mathbf{P}$;\\
3. Initialize $\mathbf{V}_0=\text{pinv}(\mathbf{U}_0)\mathbf{P}$, where $\text{pinv}(\mathbf{U}_0)$ is the\\
\ \ \ \ Moore-Penrose pseudoinverse of $\mathbf{U}_0$;\\
4. Initialize self-paced parameters $\lambda$ and $\gamma$;\\
5. \textbf{for} $t=1,\ldots, T_{max}$  \ \textbf{do} \\
6.\ \ \ \ \ \  Update $\mathbf{w}_{t}$ by solving (\ref{obj5}); \\
7.\ \ \ \ \ \  Update $\mathbf{U}_{t}$ by solving (\ref{optu}); \\
8.\ \ \ \ \ \  Update $\mathbf{V}_{t}$ by using (\ref{vv});  \\
9.\ \ \ \ \ \ $\lambda\leftarrow \lambda \mu_1$, $\gamma\leftarrow \gamma/ \mu_2$; \% \emph{update the learning pace} \\
10.\ \ \ \ \  \textbf{if}  $\|\mathbf{w}_t-\mathbf{w}_{t-1}\|_2\leq \varepsilon$ and $\|\mathbf{U}_t-\mathbf{U}_{t-1}\|_F\leq \varepsilon$ \\
\ \ \ \ \ \ \ \ \ \ \ \ and $\|\mathbf{V}_t-\mathbf{V}_{t-1}\|_F\leq \varepsilon$  \\
11.\ \ \ \  \ \ \ \ \ \textbf{break};  \\
12.\ \ \ \  \textbf{end if}\\
13. \textbf{end for} \\
\textbf{Output:} $\mathbf{w}_{t}, \mathbf{U}_t, \mathbf{V}_t.$\\
\hline
\end{tabular}
\end{center}
\end{table}
This is the standard form of system of linear equations that is full rank and thus has a unique solution.
We can solve it use the iterative methods, such as the Gauss-Seidel method \cite{courant1966methods},
which are much faster and numerically more stable than matrix inverse \cite{kumar2012learning}.

For solving $\mathbf{V}_{t+1}$ with fixed $\mathbf{w}_{t+1}$ and $\mathbf{U}_{t+1}$, the optimization problem can be decomposed into $m$ individual problems. For the $i$-th task, we have
\begin{align}
\mathbf{v}_{i,t+1}
=\arg\min\limits_{\mathbf{v}_i} \sum_{j=1}^{n_i}\frac{{w}_{j,t+1}^{(i)}}{n_i} {\mathcal{L}}(y_{ij}, \mathbf{v}_i^T\mathbf{U}_{t+1}^T\mathbf{x}_{ij})+\beta\|\mathbf{v}_i\|_1. \nonumber
\end{align}

It is hard to obtain the exact solution of the above problem directly, so we introduce an approximation scheme for efficiently solving it. It can guarantee our algorithm is convergent. The approximation is written as:
\begin{align}
\mathbf{v}_{i,t+1}=&\arg\min\limits_{\mathbf{z}} f(\mathbf{v}_{i,t}) +\nabla f(\mathbf{v}_{i,t})^T (\mathbf{z}-\mathbf{v}_{i,t})\nonumber\\
&+\frac{1}{2s_t}\|\mathbf{z}-\mathbf{v}_{i,t}\|_2^2 +h(\mathbf{z})\nonumber\\
=&\arg\min\limits_{\mathbf{z}} \frac{1}{2s_t}\|\mathbf{z}-(\mathbf{v}_{i,t}-s_t\nabla f(\mathbf{v}_{i,t})^T)\|_2^2 +h(\mathbf{z}),\nonumber
\end{align}
where $f(\mathbf{v}_{i,t})=\frac{1}{n_i}\sum_{j=1}^{n_i}{w}_{j,t+1}^{(i)}{\mathcal{L}}(y_{ij}, \mathbf{v}_{i,t}^T\mathbf{U}_{t+1}^T\mathbf{x}_{ij})$.
$\nabla f(\mathbf{v}_{i,t})$ is the derivative of $f(\mathbf{v}_i)$ around $\mathbf{v}_{i,t}$, and $h(\mathbf{z})=\beta\|\mathbf{z}\|_1$.
$s_t>0$ is a step size.
In this paper, $s_t$ is determined by a line search method \cite{beck2009gradient}.

Because of $h(\mathbf{z})=\beta\|\mathbf{z}\|_1$, we adopt the following lemma \cite{yang2009fast} to solve the above optimization problem.
\newtheorem{thm}{Lemma}
\begin{thm}\label{theo1}
For $\mu >0$, and $\mathbf{K}\in \mathbb{R}^{s\times t}$, the solution of the problem
\begin{align}
\min_{\mathbf{L}\in \mathbb{R}^{s\times t}}\mu\|\mathbf{L}\|_1+\frac{1}{2}\|\mathbf{L}-\mathbf{K}\|_F^2, \nonumber
\end{align}
is given by $\mathnormal{L}_\mu(\mathbf{K})\in\mathbb{R}^{s\times t}$, which is defined componentwisely by
\begin{align}\label{lmuk}
(\mathnormal{L}_\mu(\mathbf{K}))_{ij}:=\max\{|\mathbf{K}_{ij}|-\mu, 0\}\cdot sgn(\mathbf{K}_{ij}),
\end{align}
where $sgn(t)$ is the signum function of $t\in \mathbf{R}$.
\end{thm}

\begin{table*}
\footnotesize
\begin{center}
\caption{Results (mean$\pm$std.) on the toy dataset.
Bold font indicates that SPMTL is significantly better than the other methods  based on paired  $t$-tests at $95\%$ significance level.}
\label{toy}
\centering
\begin{tabular}{|c|c|c|c|c|c|c|c|c|c|}
\hline
 Measure &    Train           & AMTL& SPLD$\_$MTL &    GO-MTL    &  MultiSeqMT & MSMTFL & DG-MTL & SPMTL \\
\hline
\multirow{3}{*}{rMSE}  & 5\%    &5.564$\pm$0.109 &5.563$\pm$0.118&  5.745$\pm$0.018   &     5.736$\pm$0.330 &  5.704$\pm$0.056   &  5.945$\pm$0.122 & \textbf{5.447}$\pm$0.106\\
 \cline{2-9} & 10\%             &5.255$\pm$0.108&5.274$\pm$0.135&  5.731$\pm$0.101  &     5.573$\pm$0.312  &  5.566$\pm$0.117   &  5.652$\pm$0.305 & \textbf{5.075}$\pm$0.177\\
 \cline{2-9} & 15\%             &5.091$\pm$0.112&4.985$\pm$0.112&  5.179$\pm$0.244   &    5.226$\pm$0.274  &  5.376$\pm$0.070  &  5.510$\pm$0.227 & \textbf{4.694}$\pm$0.133 \\
\hline
\hline
\multirow{3}{*}{nMSE}  & 5\%    &0.943$\pm$0.018&0.964$\pm$0.022&  1.008$\pm$0.001   &     1.002$\pm$0.013  &   0.997$\pm$0.016  &  1.096$\pm$0.053 & \textbf{0.914}$\pm$0.036\\
 \cline{2-9} & 10\%             &0.834$\pm$0.026&0.938$\pm$0.019&  0.995$\pm$0.038  &      0.961$\pm$0.032  &  0.956$\pm$0.023    & 1.064$\pm$0.123 & \textbf{0.797}$\pm$0.049\\
 \cline{2-9} & 15\%             &0.786$\pm$0.048&0.845$\pm$0.064&   0.855$\pm$0.080   &    0.892$\pm$0.050  &  0.902$\pm$0.020  &   1.103$\pm$0.108 &  \textbf{0.696}$\pm$0.040\\
 \hline
\end{tabular}
\end{center}
\end{table*}

Based on the above lemma, we can obtain the solution
\begin{align}\label{vv}
(\mathbf{v}_{i,t+1})_j:=&\max\{|({\mathbf{v}_{i,t}\!-\!s_t\nabla f(\mathbf{v}_{i,t})^T})_{j}|\!-\!\beta s_t, 0\} \nonumber\\
&\cdot sgn(({{\mathbf{v}_{i,t}-s_t\nabla f(\mathbf{v}_{i,t})^T}})_j)
\end{align}

The key steps of the proposed SPMTL are summarized in Algorithm 1. In Algorithm 1, the computational complexity of updating $\mathbf{w}_t$ is of order $O(ndk)$. Updating $\mathbf{U}_t$ using Gauss-Seidel costs $O(nd^2k^2+td^2k^2)$, where $t$ denotes the number of iterations. Updating $\mathbf{V}_t$ needs $O(mdk^2)$. Therefore, the total complexity of SPTML is  $O(nd^2k^2+td^2k^2)$.
Since we utilize a convex tight upper bound to approximately solve $\mathbf{V}$, and the blocks $\mathbf{w}$ and $\mathbf{U}$ have closed-form solutions, the convergence of Algorithm 1 can be guaranteed (please see \cite{razaviyayn2013unified} for details).

\section{Experiments}
We conduct the experiments on one toy dataset and two real-world datasets to verify our method.
We compare it with several related multi-task learning (MTL) methods, including  DG-MTL \cite{kang2011learning} and GO-MTL \cite{kumar2012learning},  MultiSeqMT \cite{pentina2014curriculum}, MSMTFL  \cite{gong2013multi}, and AMTL \cite{leeasymmetric}.
In addition, we use the regularizer of SPLD instead of our proposed self-paced regularizer in (\ref{obj1}) as another baseline. we call it SPLD$\_$MTL for short.
For all the datasets, we randomly select the training instances from each task with different training ratios (5\%, 10\% and 15\%) and use the rest of instances to form the testing set.
We evaluate all the algorithms in terms of both root mean squared error (rMSE) and normalized mean squared error (nMSE).
The regularization parameter $\alpha$ in (\ref{obj4}) is used to control the complexity of the basis tasks. We find $\alpha=100$ works well on all the three datasets, and thus fix it to 100 throughout the experiments. The parameter $\beta$ is tuned in the space $[0.001, 0.01, 0.1, 1, 10, 100]$.
The parameters $\lambda$ and $\gamma$ influence how many tasks will be selected for training. Thus we initially set more than $20\%$
tasks selected in the experiment. To determine the corresponding $\lambda$ and $\gamma$, we adopt the grid search strategy based on the principle that larger $\lambda$ and smaller $\gamma$ can make more weights to be larger. After initialization, we increase $\lambda$ and decrease $\gamma$ to gradually involve hard tasks and instances at each iteration.
We repeat each case 10 times and report the average results.

\subsection{Toy Example}
\begin{table*}
\footnotesize
\begin{center}
\caption{Results (mean$\pm$std.) on the OHSUMED dataset.
Bold font indicates that SPMTL is significantly better than the other methods  based on paired  $t$-tests at $95\%$ significance level.}
\label{OHSUMED}
\centering
\begin{tabular}{|c|c|c|c|c|c|c|c|c|c|}
\hline
 Measure &    Train       &AMTL &SPLD$\_$MTL   &   GO-MTL      &  MultiSeqMT & MSMTFL & DG-MTL & SPMTL \\
\hline
\multirow{3}{*}{rMSE}  & 5\%     &0.713$\pm$0.017&0.651$\pm$0.008& 0.665$\pm$0.024   &     0.668$\pm$0.033 &  0.754$\pm$0.002  &  0.966$\pm$0.040  & \textbf{0.644}$\pm$0.007 \\
 \cline{2-9} & 10\%              &0.692$\pm$0.008&0.628$\pm$0.006&  0.651$\pm$0.018  &      0.654$\pm$0.007  & 0.756$\pm$0.003     & 0.800$\pm$0.016 & \textbf{0.624}$\pm$0.005\\
 \cline{2-9} & 15\%              &0.690$\pm$0.005&0.616$\pm$0.003& 0.626$\pm$0.010   &    0.631$\pm$0.004  &  0.788$\pm$0.006 &  0.740$\pm$0.012  & \textbf{0.614}$\pm$0.003 \\
\hline
\hline
\multirow{3}{*}{nMSE}  & 5\%     &1.482$\pm$0.121&1.243$\pm$0.111&  1.255$\pm$0.140   &     1.259$\pm$0.153 &  1.443$\pm$0.003   &  3.148$\pm$0.660 & \textbf{1.121}$\pm$0.079\\
 \cline{2-9} & 10\%              &1.312$\pm$0.036&1.109$\pm$0.023&  1.152$\pm$0.081  &      1.157$\pm$0.054  &  1.455$\pm$0.015   & 1.880$\pm$0.151 & \textbf{1.039}$\pm$0.016\\
 \cline{2-9} & 15\%              &1.296$\pm$0.058&1.073$\pm$0.034&  1.078$\pm$0.065  &    1.085$\pm$0.057  &  1.646$\pm$0.032  &  1.622$\pm$0.085  & \textbf{1.012}$\pm$0.016 \\
 \hline
\end{tabular}
\end{center}
\end{table*}

\begin{table*}
\footnotesize
\begin{center}
\caption{Results (mean$\pm$std.) on the Isolet dataset.
Bold font indicates that SPMTL is significantly better than the other methods  based on paired  $t$-tests at $95\%$ significance level.}
\label{Isolet}
\centering
\begin{tabular}{|c|c|c|c|c|c|c|c|c|c|}
\hline
 Measure &    Train     &AMTL &SPLD$\_$MTL  &   GO-MTL     &  MultiSeqMT & MSMTFL &  DG-MTL & SPMTL \\
\hline
\multirow{3}{*}{rMSE}  & 5\%    &6.374$\pm$0.382&5.930$\pm$0.167&  7.099$\pm$0.563   &    6.781$\pm$0.273 &  7.194$\pm$0.175  &   6.566$\pm$0.228  & \textbf{5.909}$\pm$0.142 \\
 \cline{2-9} & 10\%             &5.902$\pm$0.067&5.605$\pm$0.034 & 6.189$\pm$0.267  &     6.104$\pm$0.184  &  6.494$\pm$0.105  & 6.168$\pm$0.166 & \textbf{5.570}$\pm$0.036\\
 \cline{2-9} & 15\%             &5.880$\pm$0.043&5.468$\pm$0.051 & 5.519$\pm$0.124  &     5.833$\pm$0.079  &  6.173$\pm$0.085 &  6.043$\pm$0.098  &  \textbf{5.444}$\pm$0.059 \\
\hline
\hline
\multirow{3}{*}{nMSE}  & 5\%    &0.724$\pm$0.028&0.803$\pm$0.032&   0.905$\pm$0.146  &    0.772$\pm$0.083 &   0.921$\pm$0.045  &  0.768$\pm$0.055 & \textbf{0.621}$\pm$0.030\\
 \cline{2-9} & 10\%             &0.621$\pm$0.064&0.729$\pm$0.026&  0.688$\pm$0.059  &     0.669$\pm$0.047  &     0.751$\pm$0.025    & 0.678$\pm$0.038 & \textbf{0.552}$\pm$0.008\\
 \cline{2-9} & 15\%             &0.619$\pm$0.017&0.717$\pm$0.016 &  0.541$\pm$0.024   &    0.557$\pm$0.023 &  0.677$\pm$0.018  &  0.650$\pm$0.021 & \textbf{0.526}$\pm$0.012 \\
 \hline
\end{tabular}
\end{center}
\end{table*}

We first describe the synthetic data generation procedure.
Let there be 3 groups and each group has 10 tasks.
There are 100 instances in each task; each instance is represented by a 15-dimensional vector.
We generate parameter vectors for 4 latent tasks, i.e., $\mathbf{U}$ in the proposed formulation, in 20 dimensions, with each entry drawn i.i.d. from a standard normal distribution.
Based on $\mathbf{U}$, we generate the first 10 tasks by linearly combining only the first two latent tasks.
In a similar manner, the next 10 tasks are generated by linearly combining the second and the third latent tasks.
Last 10 task are generated by linear combinations of the last two latent tasks.
All the coefficients of linear combinations, i.e., $\mathbf{V}$, are drawn i.i.d. from a standard normal distribution.
The instance $\mathbf{x}_{ij}$ is sampled from a standard Gaussian distribution, and the response is  $y_{ij}=\mathbf{v}_i^T\mathbf{U}^T\mathbf{x}_{ij}+\xi_{ij}$. To create hard tasks, we add different noise to tasks and instances by
setting $\xi_{ij}=\sigma_i\theta_j$, where $\sigma_i$'s are i.i.d. from a normal distribution $N(0,5)$, and $\theta_j$ is drawn i.i.d. from $N(0,1)$.

We first report the statistical results on this dataset as shown in Table \ref{toy}.
SPMTL achieves the best result among all the methods under different training ratios.
This means that incorporating the easy-to-hard strategy on both instance level and task level into the learning process can improve the prediction performance.
Moreover, SPMTL is better than GO-MTL, a task grouping method, and  MultiSeqMT, a task selection method. It indicates that only learning related information among a subset of tasks without task selection, or only selecting tasks without learning grouping information is not optimal for MTL.
Finally, SPMTL significantly outperforms SPLD$\_$MTL, which shows our task-oriented self-paced regularizer is better for MTL than that of SPLD.
\begin{figure}
\centering
\subfigure[rMSE]{\includegraphics[width=0.3\linewidth]{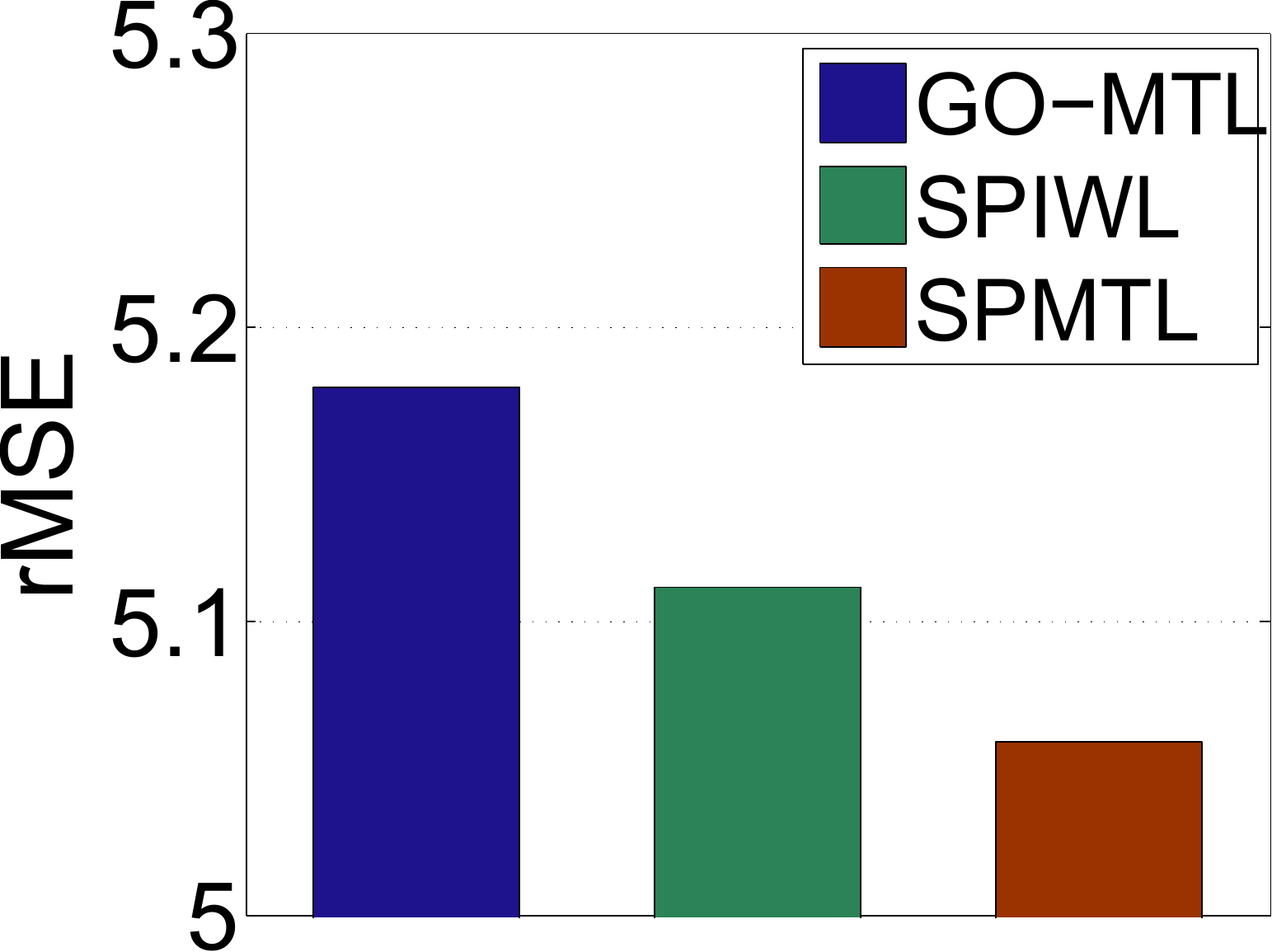}}
\subfigure[nMSE]{\includegraphics[width=0.3\linewidth]{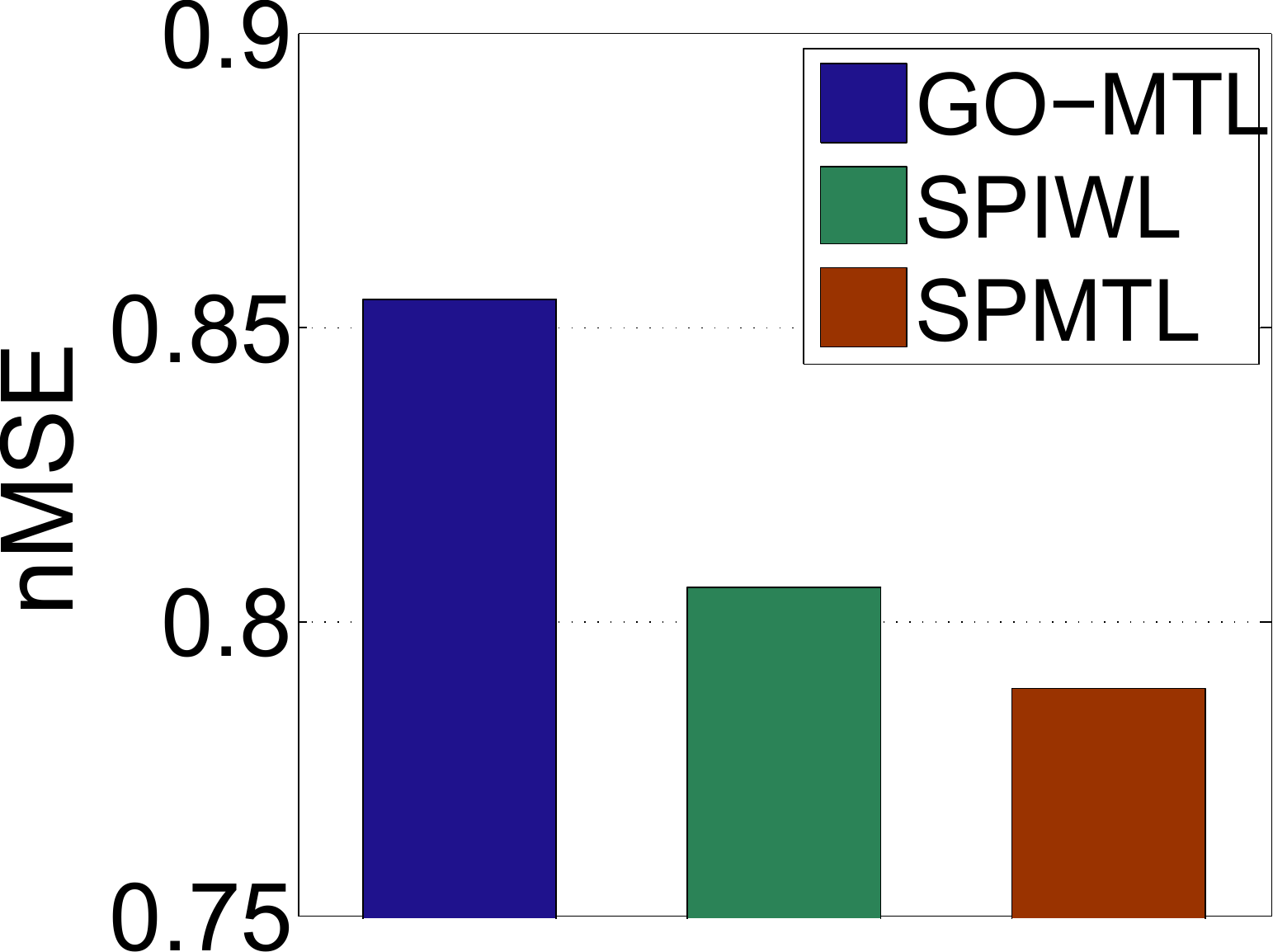}}
\caption{Effectiveness verification of considering the order of both tasks and instances on the toy dataset.}
\label{fig:component}
\end{figure}

We also test the effectiveness of considering either or both instance order and task order in our method. By setting $\gamma=0$ in (\ref{obj4}), we only consider the complexities of the instances. We call it Self-Paced Instance Weight Learning (SPIWL).
The experiments are conducted on the $15\%$ training data, and the results are shown in Figure \ref{fig:component}.
Since GO-MTL is our special case (when $\lambda=\gamma=0$, and $\mathbf{w}=1$, our method is reduced to GO-MTL), we take it as the baseline.
SPIWL performs better than GO-MTL. This suggests that including the instances from the easiest to the hardest improves the performance.
SPMTL outperforms SPIWL, which demonstrates that involving the tasks based on the easy-to-hard strategy can also be helpful for model training.

\subsection{Real-World Data Experiments}
\begin{figure}
\centering
{\includegraphics[width=0.6\linewidth]{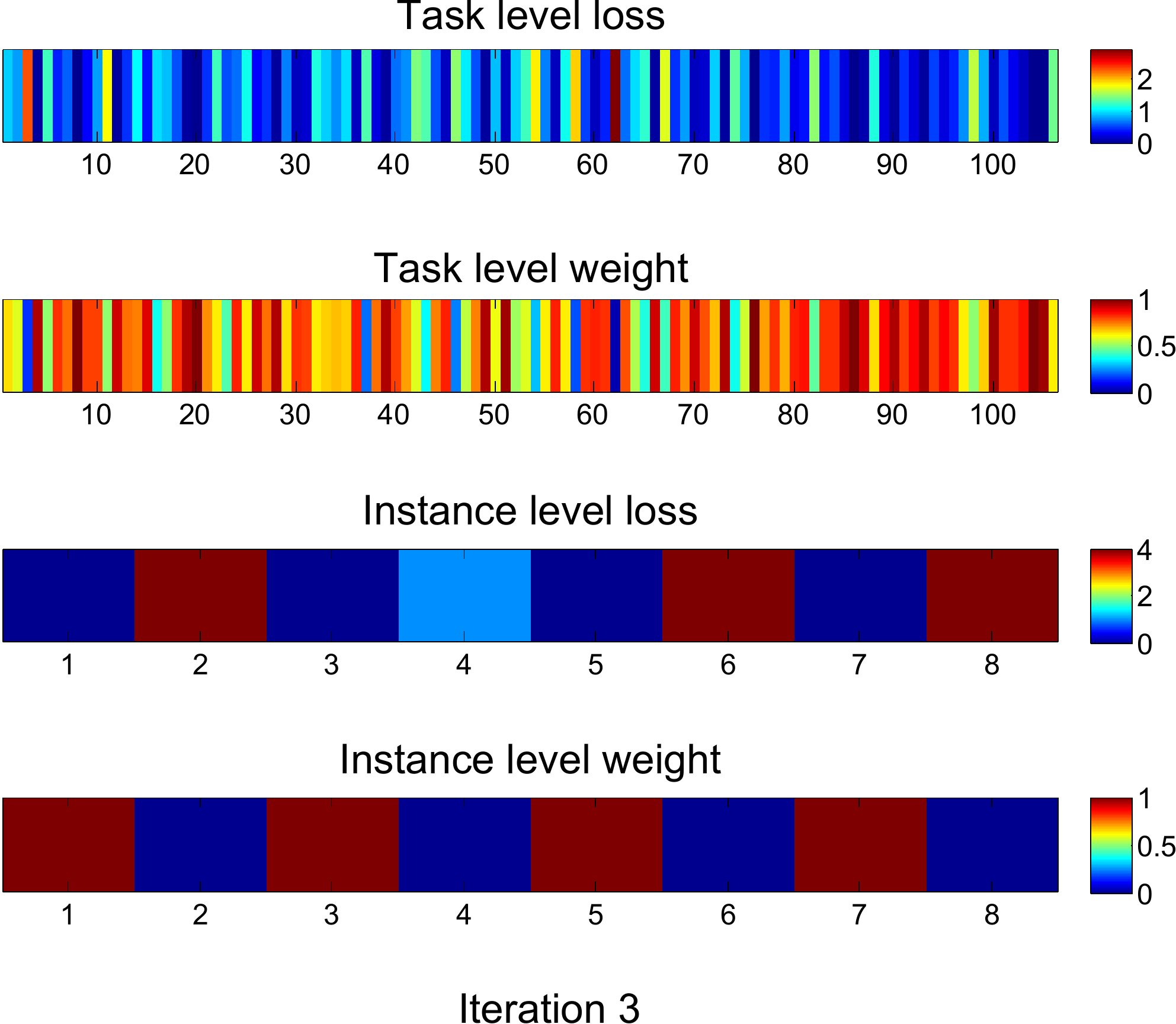}}
\caption{
An example on tasks and instances selected by Algorithm 1. Dark blue denotes the values are close to zero.
}
\label{ww}
\end{figure}
In this section, we conduct the experiments on two real-world datasets: OHSUMED \cite{hersh1994ohsumed} and Isolet\footnote{http://www.cad.zju.edu.cn/home/dengcai/Data/MLData.html}.
The first one is an ordinal regression dataset which consists of 106 queries. We take each query as one task. Each query comes with multiple returned documents with labels indicating how relevant the returned document is to the query: ``definitely relevant'', ``possibly relevant'', or ``not relevant''.  These documents and their relevance labels are the instances to the corresponding query (task).
Each query is associated with 70 instances in average, and there are in total 7,546 instances with the feature dimension of 25.
The second dataset is collected from 150 speakers who speak  each English letter of the alphabet twice.
Thus there are 52 samples from each individual.
Each English letter corresponds to a label (1-26), and the label is treated as the regression value as in \cite{gong2013multi}.
The individuals are grouped into 5 groups by speaking similarity. Thus, we naturally have 5 tasks with each task corresponding to a group.
There are 1560, 1560, 1560, 1558, and 1559 instances in the 5 tasks respectively. Each instance is represented by a 617-dimensional vector. We reduce dimensions using PCA with 90\% of the variance retaining, in order to learn efficiently.

Tables \ref{OHSUMED} and \ref{Isolet} report the performance measured by rMSE and nMSE for the OHSUMED dataset and the Isolet dataset, respectively.
Our SPMTL significantly outperforms all the other methods on both datasets.
This demonstrates that our method is effective by incorporating the self-paced learning regime into MTL once more.
\begin{figure}
\centering
\subfigure{\includegraphics[width=0.45\linewidth]{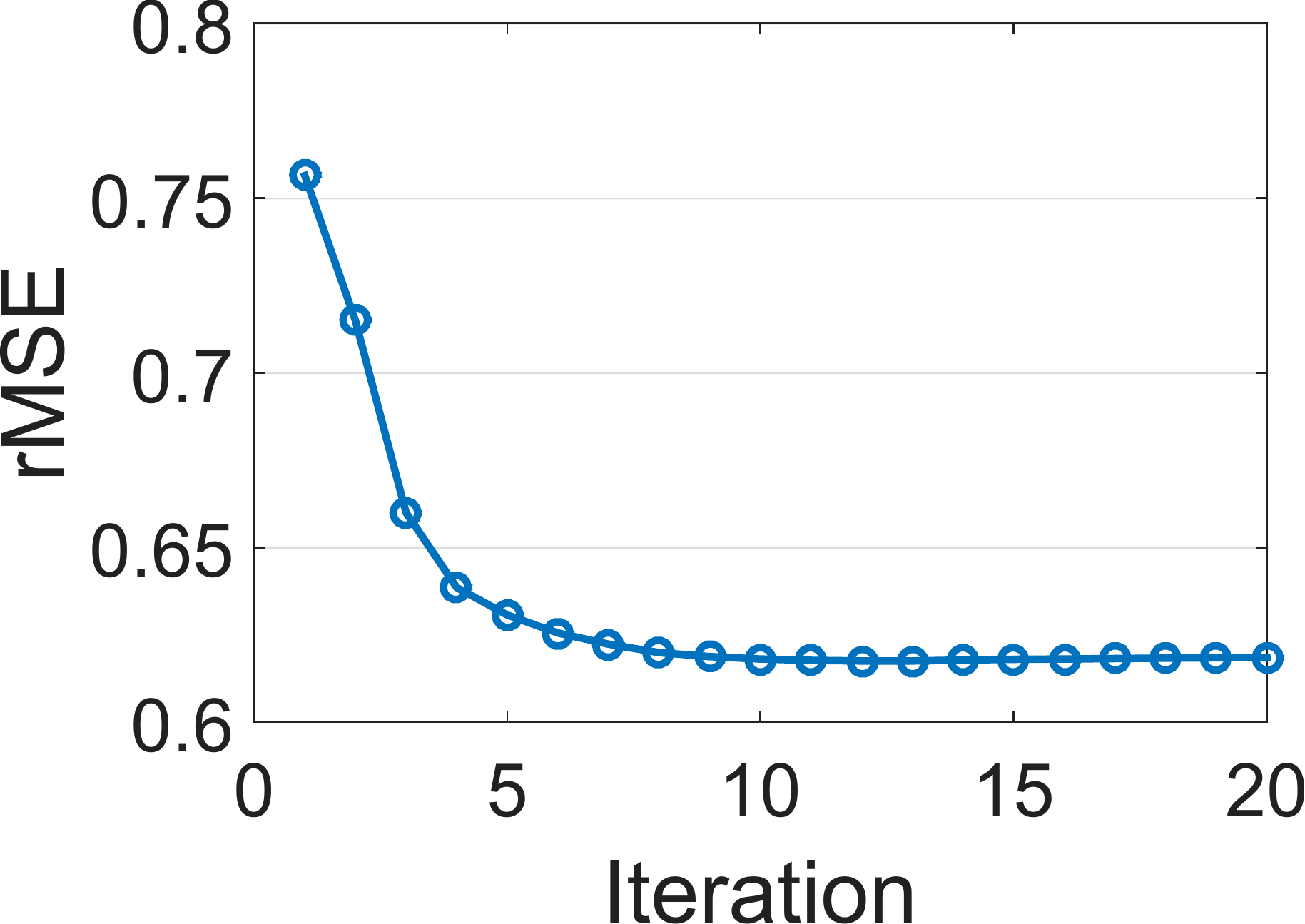}}
\subfigure{\includegraphics[width=0.45\linewidth]{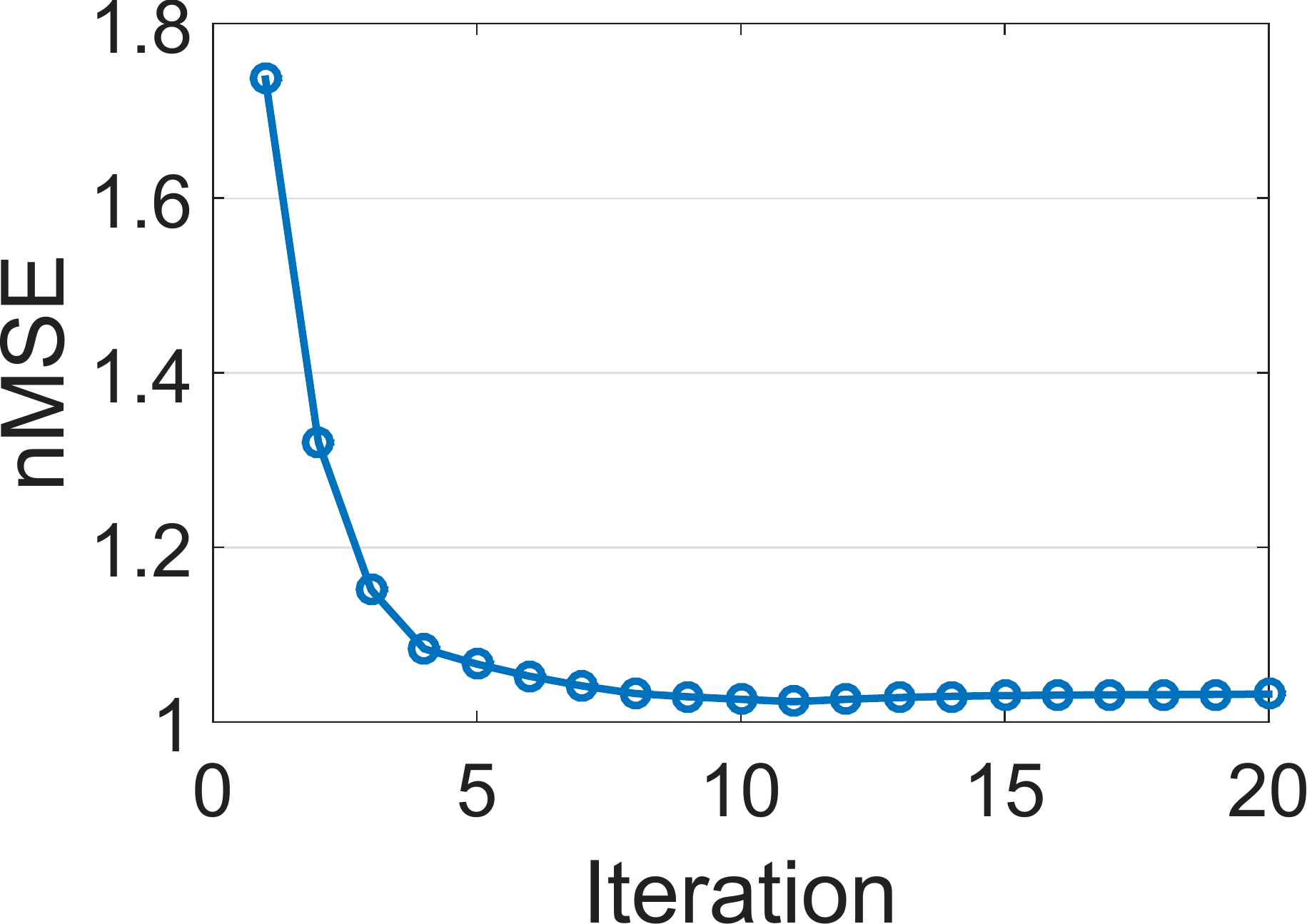}}
\caption{Prediction performance vs. iterations.}
\label{fig:convergence}
\end{figure}

We visualize $\mathbf{w}$ and $\mathcal{L}$ in (\ref{obj4}) using $15\%$ training data on the OHSUMED dataset, as shown in Figure \ref{ww}.
The first two pictures depict the averaged loss and averaged weight of each task.  When the loss is small (easy task), the corresponding weight is large, thus the easy tasks are first selected for training. The third and fourth pictures show the loss and weight on the instance level in the $i$-th task (here $i=10$).
Similarly, the instances with lower loss have higher weight, i.e., easy instances are first selected for training.

Finally, we further study the prediction performance of SPMTL as the iteration increases on the OHSUMED dataset with $15\%$ training data.
The results are shown in Figure \ref{fig:convergence}.
Only after around 10 iterations, the performances of SPMTL become stable, which implies SPMTL is then convergent.

\section{Conclusion and Future Work}
We present a novel multi-task learning algorithm, namely SPMTL. We incorporate the easy-to-hard strategy on both tasks and instances into the learning process of multi-task learning. Experiments on both synthetic dataset and real datasets have verified the effectiveness of SPMTL.

A question is there should be more complicated patterns for instance selection in the context of multi-task learning.
For example, prioritizing easy instances in difficult tasks and difficult instances in
easy tasks can be helpful for model training. We will study it in future work especially in the context of joint learning \cite{LiArxiv15}.
\section{ Acknowledgments}
The work was supported by the IBM Shared Unison Research Program 2015-2016, Natural Science Foundation of China (Grant No. 61532009, 61602176, 61672231), China Postdoctoral Science Foundation Funded Project (Grant No. 2016M590337), the Funding of Jiangsu Province (Grant No. 15KJA520001) and NSF (IIS-1639792, DMS-1620345). We sincerely thank Dr. Xiangfeng Wang for his valuable suggestions to improve this work.
\small
\bibliographystyle{aaai}
\bibliography{sigproc}
\end{document}